\def\year{2020}\relax
\documentclass[letterpaper]{article} 
\usepackage{aaai20}  
\usepackage{times}  
\usepackage{helvet} 
\usepackage{courier}  
\usepackage[hyphens]{url}  
\usepackage{graphicx} 
\usepackage{graphicx}  
\usepackage{makecell}
\usepackage{multirow}
\usepackage{xcolor}
\usepackage{amssymb}
\usepackage{subcaption,siunitx,booktabs}

\usepackage[protrusion=true,expansion=true]{microtype}

\title{MMM: Multi-stage Multi-task Learning for Multi-choice Reading Comprehension}

\author{Di Jin\textsuperscript{\rm 1}\thanks{This work was done when the author was an intern at Amazon Alexa AI, Sunnyvale, CA.}, Shuyang Gao\textsuperscript{\rm 2}, Jiun-Yu Kao\textsuperscript{\rm 2}, Tagyoung Chung\textsuperscript{\rm 2}, Dilek Hakkani-tur\textsuperscript{\rm 2}\\  
\textsuperscript{\rm 1}Computer Science \& Artificial Intelligence Laboratory, MIT, MA, USA\\
\textsuperscript{\rm 2}Amazon Alexa AI, Sunnyvale, CA, USA \\
jindi15@mit.edu, \{shuyag,jiunyk,tagyoung,hakkanit\}@amazon.com 
}

\begin{document}

\maketitle

\begin{abstract}

Machine Reading Comprehension (MRC) for question answering (QA), which aims to answer a question given the relevant context passages, is an important way to test the ability of intelligence systems to understand human language.
Multiple-Choice QA (MCQA) is one of the most difficult tasks in MRC because it often requires more advanced reading comprehension skills such as logical reasoning, summarization, and arithmetic operations, compared to the extractive counterpart where answers are usually spans of text within given passages. Moreover, most existing MCQA datasets are small in size, making the learning task even harder.
We introduce \textbf{{MMM}}, a \textbf{M}ulti-stage \textbf{M}ulti-task learning framework for \textbf{M}ulti-choice reading comprehension. Our method involves two sequential stages: coarse-tuning stage using out-of-domain datasets and multi-task learning stage using a larger in-domain dataset to help model generalize better with limited data. Furthermore, we propose a novel multi-step attention network (MAN) as the top-level classifier for this task. We demonstrate MMM significantly advances the state-of-the-art on four representative MCQA datasets.\footnote{Code is released: https://github.com/jind11/MMM-MCQA}
\end{abstract}

\section{Introduction}

Building a system that comprehends text and answers questions is challenging but fascinating, which can be used to test the machine's ability to understand human language~\cite{hermann2015teaching,chen2018neural}. 
Many machine reading comprehension (MRC) based question answering (QA) scenarios and datasets have been introduced over the past few years, which differ from each other in various ways, including the source and format of the context documents, whether external knowledge is needed, the format of the answer, to name a few. We can divide these QA tasks into two categories: 1) extractive/abstractive QA such as SQuAD \cite{rajpurkar2018know}, and HotPotQA \cite{yang2018hotpotqa}. 
2) multiple-choice QA (MCQA) tasks such as MultiRC \cite{khashabi2018looking}, and MCTest \cite{ostermann2018semeval}. 

In comparison to extractive/abstractive QA tasks, the answers of the MCQA datasets are in the form of open, natural language sentences and not restricted to spans in text. Various question types exist such as arithmetic, summarization, common sense, logical reasoning, language inference, and sentiment analysis. Therefore it  requires more advanced reading skills for the machine to perform well on this task. Table~\ref{table: example} shows one example from one of MCQA datasets, DREAM \cite{sun2019dream}. To answer the first question in Table~\ref{table: example}, the system needs to comprehend the whole dialogue and use some common sense knowledge to infer that such a conversation can only happen between classmates rather than brother and sister. For the second question, the implicit \textit{inference} relationship between the utterance \textit{``You'll forget your head if you're not careful.''} in the passage and the answer option \textit{``He is too careless.''} must be figured out by the model to obtain the correct answer. Many MCQA datasets were collected from language or science exams, which were purposely designed by educational experts and consequently require non-trivial reasoning techniques \cite{lai2017race}. As a result, the performance of machine readers on these tasks can more accurately gauge comprehension ability of a model. 

\begin{table}
\centering
\small
\resizebox{\columnwidth}{!}{\begin{tabular}{l}
\Xhline{2\arrayrulewidth}
\textbf{Dialogue}                                                               \\ \hline
W: Come on, Peter! It's nearly seven.                                \\ 
M: I'm almost ready. \\ 
W: We'll be late if you don't hurry.             \\ 
M: One minute, please. I'm packing my things. \\ 
W: The teachers won't let us in if we are late. \\ 
M: Ok. I'm ready. Oh, I'll have to get my money. \\
W: You don't need money when you are having the exam, do you? \\
M: Of course not. Ok, let's go... Oh, my god. I've forgot my watch. \\
W: \textbf{\textit{You'll forget your head if you're not careful.}}\\
M: My mother says that, too. \\ \hline
\textbf{Question 1}: What's the relationship between the speakers?                                       \\ 
\textbf{A.} Brother and sister.\hspace{2.1 mm}   \textbf{B.} Mother and son.  \hspace{7.3 mm} \textbf{C.} Classmates. $\surd$ \\ \hline
\textbf{Question 2}: What does the woman think of the man? \\ 
\textbf{A.} He is very serious. \hspace{1 mm} \textbf{B.} \textbf{\textit{He is too careless.}} $\surd$ \hspace{1 mm} \textbf{C.} He is very lazy.                                              \\ \Xhline{2\arrayrulewidth}
\end{tabular}}
\caption{\small Data samples of DREAM dataset. ($\surd$: the correct answer)}
\label{table: example}
\end{table}

Recently large and powerful pre-trained language models such as BERT \cite{devlin-etal-2019-bert} have been achieving the state-of-the-art (SOTA) results on various tasks, however, its potency on MCQA datasets has been severely limited by the data insufficiency.
For example, the MCTest dataset has two variants: MC160 and MC500, which are curated in a similar way, and MC160 is considered easier than MC500 \cite{richardson2013mctest}. However, BERT-based models perform much worse on MC160 compared with MC500 (8--10\% gap) since the data size of the former is about three times smaller.
To tackle this issue, we investigate how to improve the generalization of BERT-based MCQA models with the constraint of limited training data using four representative MCQA datasets: DREAM, MCTest, TOEFL, and SemEval-2018 Task 11\@. 

We proposed \textbf{MMM}, a \textbf{M}ulti-stage \textbf{M}ulti-task  learning framework for \textbf{M}ulti-choice question answering. Our framework involves two sequential stages: coarse-tuning stage using out-of-domain datasets and multi-task learning stage using a larger in-domain dataset. For the first stage, we coarse-tuned our model with natural language inference (NLI) tasks. For the second multi-task fine-tuning stage, we leveraged the current largest MCQA dataset, RACE, as the in-domain source dataset and simultaneously fine-tuned the model on both source and target datasets via multi-task learning. Through extensive experiments, we demonstrate that the two-stage sequential fine-tuning strategy is the optimal choice for BERT-based model on MCQA datasets. Moreover, we also proposed a Multi-step Attention Network (MAN) as the top-level classifier instead of the typical fully-connected neural network for this task and obtained better performance. Our proposed method improves BERT-based baseline models by at least 7\% in absolute accuracy for all the MCQA datasets (except the SemEval dataset that already achieves 88.1\% for the baseline). As a result, by leveraging BERT and its variant, RoBERTa~\cite{liu2019roberta}, our approach advanced the SOTA results for all the MCQA datasets, surpassing the previous SOTA by at least 16\% in absolute accuracy (except the SemEval dataset).

\section{Methods}


In MCQA, the inputs to the model are a passage, a question, and answer options. The passage, denoted as $P$, consists of a list of sentences. The question and each of the answer options, denoted by $Q$ and $O$, are both single sentences. A MCQA model aims to choose one correct answer from answer options based on $P$ and $Q$.

\subsection{Model Architecture}

\begin{figure}
    \centering
    \includegraphics[width=0.47\textwidth]{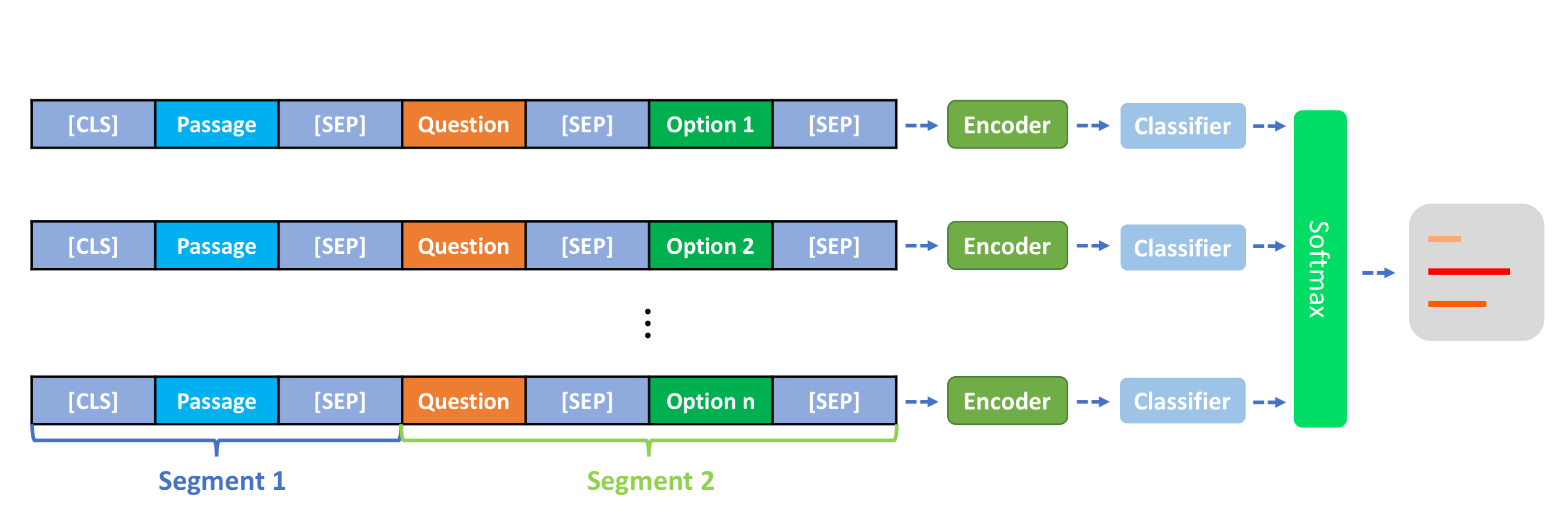}
    \caption{\small Model architecture. ``Encoder''is a pre-trained sentence encoder such as BERT. ``Classifier'' is a top-level classifier.}
    \label{fig:model}
\end{figure}

Figure~\ref{fig:model} illustrates the model architecture. Specifically, we concatenate the passage, question and one of the answer options into a long sequence. For a question with $n$ answer options, we obtain $n$ token sequences of length $l$. Afterwards, each sequence will be encoded by a sentence encoder to get the representation vector $H \in \mathbb{R}^{d\times l}$, which is then projected into a single value $p=C(H)$ ($p\in \mathbb{R}^{1}$) via a top-level classifier $C$. In this way, we obtain the logit vector $\mathbf{p}=[p_1,p_2,...,p_n]$ for all options of a question, which is then transformed into the probability vector through a softmax layer. We choose the option with highest logit value $p$ as the answer. Cross entropy loss is used as the loss function. We used the pre-trained bidirectional transformer encoder, i.e., BERT and RoBERTa as the sentence encoder. The top-level classifier will be detailed in the next subsection.

\subsection{Multi-step Attention Network}
\label{section:man}

For the top-level classifier upon the sentence encoder, the simplest choice is a two-layer full-connected neural network (FCNN), which consist of one hidden layer with $tanh$
activation and one output layer without activation. This has been widely adopted when BERT is fine-tuned for the down-streaming classification tasks and performs very well \cite{devlin-etal-2019-bert}. Inspired from the success of the attention network widely used in the span-based QA task \cite{seo2016bidirectional}, we propose the multi-step attention network (MAN) as our top-level classifier. Similar to the dynamic or multi-hop memory network \cite{kumar2016ask,liu2017stochastic}, MAN maintains a state and iteratively refines its prediction via the multi-step reasoning.  

The MAN classifier works as follows. A pair of question and answer option together is considered as a whole segment, denoted as $QO$. Suppose the sequence length of the passage is $p$ and that of the question and option pair is $q$. We first construct the working memory of the passage $H^P\in \mathbb{R}^{d\times p}$ by extracting the hidden state vectors of the tokens that belong to $P$ from $H$ and concatenating them together in the original sequence order. Similarly, we obtain the working memory of the (question, option) pair, denoted as $H^{QO}\in \mathbb{R}^{d\times q}$. Alternatively, we can also encode the passage and (question, option) pair individually to get their representation vectors $H^P$ and $H^{QO}$, but we found that processing them in a pair performs better.

We then perform $K$-step reasoning over the memory to output the final prediction. Initially, the initial state $\mathbf{s}^0$ in step 0 is the summary of $H^P$ via self-attention: $\mathbf{s}^0=\sum_i \alpha_i H_i^P$, where $\alpha_i=\frac{exp(w_1^TH_i^P)}{\sum_j exp(w_1^TH_j^P)}$. In the following steps $k \in {1,2,...,K-1}$, the state is calculated by:
\begin{equation}
    \mathbf{s}^k=GRU(\mathbf{s}^{k-1}, \mathbf{x}^k),
\end{equation}
where $\mathbf{x}^k=\sum_i\beta_iH_i^{QO}$ and $\beta_i=\frac{exp(w_2^T[\mathbf{s}^{k-1};H_i^{QO}])}{\sum_j exp(w_2^T[\mathbf{s}^{k-1};H_j^{QO}])}$. Here $[x;y]$ is concatenation of the vectors $x$ and $y$. The final logit value is determined using the last step state:
\begin{equation}
    P=w_3^T[\mathbf{s}^{K-1};\mathbf{x}^{K-1};|\mathbf{s}^{K-1}-\mathbf{x}^{K-1}|;\mathbf{s}^{K-1}\cdot\mathbf{x}^{K-1}].
\end{equation}
Basically, the MAN classifier calculates the attention scores between the passage and (question, option) pair step by step dynamically such that the attention can refine itself through several steps of deliberation. The attention mechanism can help filter out irrelevant information in the passage against (question, option) pair.

\subsection{Two Stage Training}
\label{section:fine-tuning}

\begin{figure}
    \centering
    \includegraphics[width=0.34\textwidth]{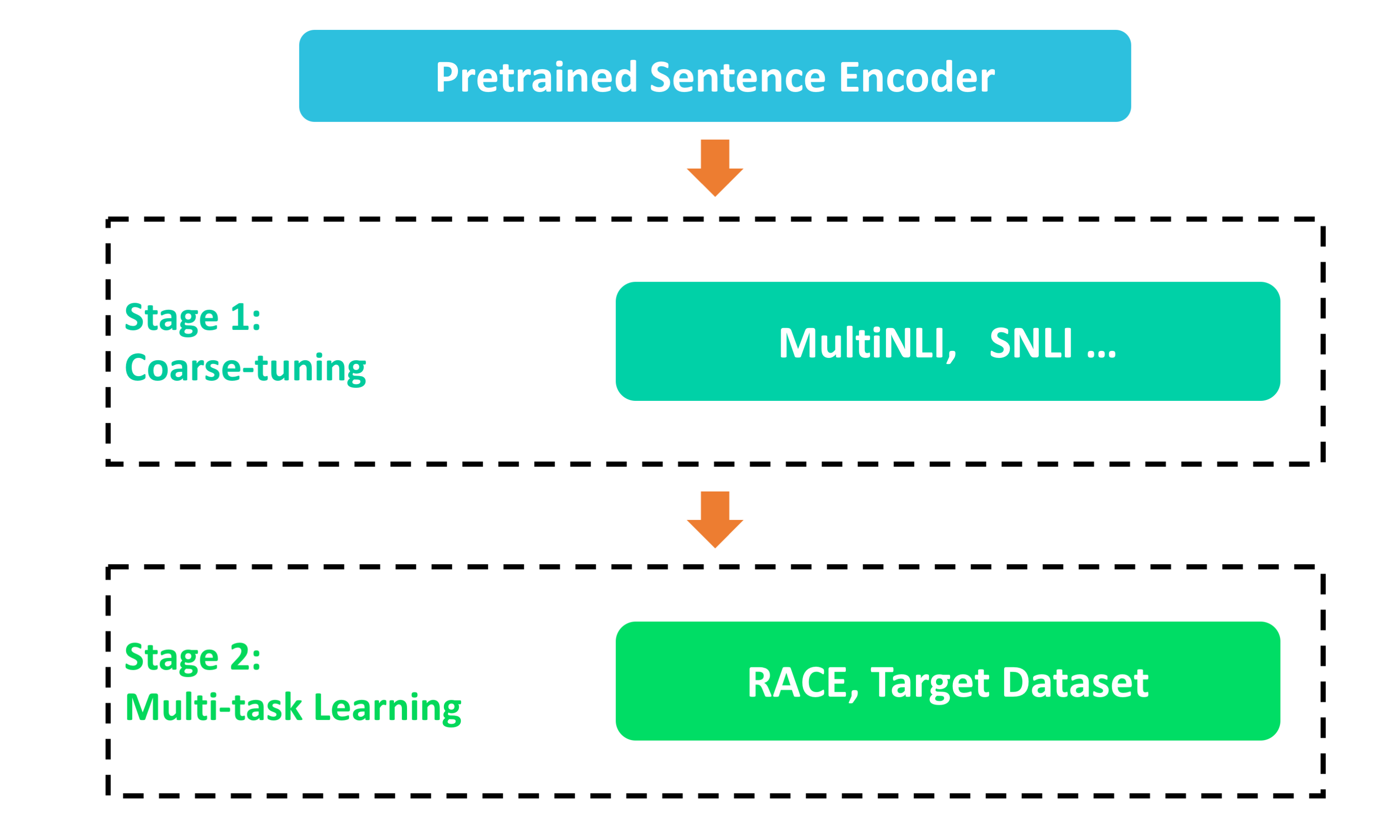}
    \caption{\small Multi-stage and multi-task fine-tuning strategy.}
    \label{fig:tuning}
\end{figure}

We adopt a two-stage procedure to train our model with both in-domain and out-of-domain datasets as shown in Figure~\ref{fig:tuning}. 

\paragraph{Coarse-tuning Stage} We first fine-tune the sentence encoder of our model with natural language inference (NLI) tasks. For exploration, we have also tried to fine-tune the sentence encoder on other types of tasks such as sentiment analysis, paraphrasing, and span-based question answering  at this stage. However, we found that only NLI task shows robust and significant improvements for our target multi-choice task. See Section~\ref{sec:discussion} for details.

\paragraph{Multi-task Learning Stage} After corase-tuning stage, we simultaneously fine-tune our model on a large in-domain source dataset and the target dataset together via multi-task learning. We share all model parameters including the sentence encoder as well as the top-level classifier for these two datasets. 

\begin{table*}
\centering
\small
\begin{tabular}{lrrrrr}
\Xhline{2\arrayrulewidth}
                            & \textbf{DREAM}     & \textbf{MCTest}          & \textbf{SemEval-2018 Task 11} & \textbf{TOEFL}          & \textbf{RACE}         \\ \hline
construction method         & exams     & crowd.          & crowd.               & exams          & exams        \\
passage type                & dialogues & child's stories & narrative text       & narrative text & written text \\ 
\# of options               & 3         & 4               & 2                    & 4              & 4            \\ 
\# of passages              & 6,444     & 660             & 2,119                & 198            & 27,933       \\ 
\# of questions             & 10,197    & 2,640           & 13,939               & 963            & 97,687       \\ \hline
non-extractive answer$^\star$ (\%) & 83.7      & 45.3            & 89.9                 & -              & 87.0         \\ \Xhline{2\arrayrulewidth}
\end{tabular}
\caption{\small Statistics of MCQA datasets. (crowd.: crowd-sourcing; $\star$: answer options are not text snippets from reference documents.)}
\label{table:datasets}
\end{table*}

\section{Experimental Setup}

\subsection{Datasets}

We use four MCQA datasets as the target datasets: DREAM \cite{sun2019dream}, MCTest \cite{richardson2013mctest}, TOEFL \cite{ostermann2018semeval}, and SemEval-2018 Task 11 \cite{tseng2016towards}, which are summarized in Table~\ref{table:datasets}. For the first coarse-tuning stage with NLI tasks, we use MultiNLI
\cite{williams2017broad} and SNLI
\cite{young2014image} as the out-of-domain source datasets. For the second stage, we use the current largest MCQA dataset, i.e., RACE \cite{lai2017race} as in-domain source dataset. For all datasets, we use the official train/dev/test splits.

\subsection{Speaker Normalization}

Passages in DREAM dataset are dialogues between two persons or more. Every utterance in a dialogue starts with the speaker name. For example, in utterance ``m: How would he know?'', ``m'' is the abbreviation of ``man'' indicating that this utterance is from a man. More than 90\% utterances have the speaker names as ``w,'' ``f,'' and ``m,'' which are all abbreviations. However, the speaker names mentioned in the questions are full names such as ``woman'' and ``man.'' In order to make it clear for the model to learn which speaker the question is asking about, we used a speaker normalization strategy by replacing ``w'' or ``f'' with ``woman'' and ``m'' with ``man'' for the speaker names in the utterances. We found this simple strategy is quite effective, providing us with 1\% improvement. We will always use this strategy for the DREAM dataset for our method unless explicitly mentioned.

\subsection{Multi-task Learning}

For the multi-task learning stage, at each training step, we randomly selected a dataset from the two datasets (RACE and the target dataset) and then randomly fetched a batch of data from that dataset to train the model. This process was repeated until the predefined maximum number of steps or the early stopping criterion has been met. We adopted the proportional sampling strategy, where the probability of sampling a task is proportional to the relative size of each dataset compared to the cumulative size of all datasets \cite{liu2019multi}.

\subsection{Training Details}

We used a linear learning rate decay schedule with warm-up proportion of $0.1$. We set the dropout rate as $0.1$\@. The maximum sequence length is set to 512. We clipped the gradient norm to $5$ for DREAM dataset and $0$ for other datasets. The learning rate and number of training epochs vary for different datasets and encoder types, which are summarized in the Appendix \ref{appendix:section-1}. The model architecture and training settings for the NLI task are the same as those in \cite{devlin-etal-2019-bert}.

More than 90\% of passages have more than $512$ words in the TOEFL dataset, which exceed the maximum sequence length that BERT supports, thus we cannot process the whole passage within one forward pass. To solve this issue, we propose the sliding window strategy, in which we split the long passage into several snippets of length 512 with overlaps between subsequent snippets and each snippet from the same passage will be assigned with the same label. In training phase, all snippets will be used for training, and in inference phase, we aggregate the logit vectors of all snippets from the same passage and pick the option with highest logit value as the prediction. In experiments, we found the overlap of 256 words is the optimal, which can improve the BERT-Base model from accuracy of 50.0\% to 53.2\%. We adopted this sliding window strategy \textbf{only} for the TOEFL dataset.

\section{Results}

\begin{table}
\centering
\small
\begin{tabular}{lrr}
\Xhline{2\arrayrulewidth}
\textbf{Model}               & \textbf{Dev}  & \textbf{Test} \\ \hline
FTLM++ \cite{sun2019dream}   & 58.1 & 58.2 \\ 
BERT-Large \cite{devlin-etal-2019-bert}          & 66.0 & 66.8 \\ 
XLNet \cite{yang2019xlnet}               & -    & 72.0 \\ \hline
BERT-Base           &  63.2    &    63.2  \\ 
BERT-Large          &  66.2    &  66.9    \\ 
RoBERTa-Large       &  85.4    &  85.0    \\
\hline
BERT-Base+MMM           &   72.6 (9.4)   & 72.2 (9.0)     \\ 
BERT-Large+MMM          &  75.5 (9.3)    & 76.0 (9.1)     \\ 
RoBERTa-Large+MMM       &  \textbf{88.0} (2.6)    & \textbf{88.9} (3.9)     \\ 
\hline\hline
Human Performance   & 93.9$^\star$ & 95.5$^\star$ \\ 
Ceiling Performance & 98.7$^\star$ & 98.6$^\star$ \\ \Xhline{2\arrayrulewidth}
\end{tabular}
\caption{\small Accuracy on the DREAM dataset. Performance marked by $\star$ is reported by \cite{sun2019dream}. Numbers in parentheses indicate the accuracy increased by MMM compared to the baselines.}
\label{table:results-dream}
\end{table}

\begin{table*}
\centering
\small
\resizebox{1.\textwidth}{!}{\begin{tabular}{llccccccc}
\Xhline{2\arrayrulewidth}
\multirow{2}{*}{\textbf{Dataset}} &  \multirow{2}{*}{\textbf{Previous Single-Model SOTA}} & \multicolumn{3}{c}{\textbf{Baselines}} & \multicolumn{3}{c}{\textbf{+MMM}} & \textbf{Human}\\
      &  & \textbf{BERT-B} & \textbf{BERT-L} & \textbf{RoBERTa-L} & \textbf{BERT-B} & \textbf{BERT-L} & \textbf{RoBERTa-L} & \textbf{Scores} \\ \hline
MC160 & 80.0 \cite{sun2018improving}                       &   63.8        &   65.0         &   81.7      &   85.4 (21.6) &   89.1 (\textbf{24.1})  & \textbf{97.1} (15.4) & 97.7$^\star$ \\ 
MC500 & 78.7 \cite{sun2018improving}                      &  71.3         &   75.2        &  90.5       &  82.7 (11.4) & 86.0 (10.8) & \textbf{95.3} (4.8) & 96.9$^\star$       \\ 
TOEFL        & 56.1 \cite{chung2017supervised}                  &  53.2         &     55.7       &    64.7     &  60.7 (7.5)   & 66.4 (10.7)  & \textbf{82.8} (18.1)   &   --   \\ 
SemEval      & 88.8 \cite{sun2018improving}                      &  88.1         &    88.7        &   94.0      &    89.9 (1.8)  & 91.0 (2.3)  &  \textbf{95.8} (1.8)  & 98.0$^\dagger$    \\ \Xhline{2\arrayrulewidth}
\end{tabular}
}
\caption{\small Performance in accuracy (\%) on test sets of other datasets: MCTest (MC160 and MC500), TOEFL, and SemEval. Performance marked by $\star$ is reported by \cite{richardson2013mctest} and that marked by $\dagger$ is from \cite{ostermann2018semeval}. Numbers in the parentheses indicate the accuracy increased by MMM. ``-B'' means the base model and ``-L'' means the large model.}
\label{table:results-other-datasets}
\end{table*}

We first evaluate our method on the DREAM dataset. The results are summarized in Table~\ref{table:results-dream}. In the table, we first report the accuracy of the SOTA models in the leaderboard. We then report the performance of our re-implementation of fine-tuned models as another set of strong baselines, among which the RoBERTa-Large model has already surpassed the previous SOTA. For these baselines, the top-level classifier is a two-layer FCNN for BERT-based models and a one-layer FCNN for the RoBERTa-Large model. Lastly, we report model performances that use all our proposed method, \textbf{\textit{MMM}} (MAN classifier + speaker normalization + two stage learning strategies). As direct comparisons, we also list the accuracy increment between MMM and the baseline with the same sentence encoder marked by the parentheses, from which we can see that the performance augmentation is over 9\% for BERT-Base and BERT-Large. Although the RoBERTa-Large baseline has already outperformed the BERT-Large baseline by around 18\%, MMM gives us another $\sim$4\% improvement, pushing the accuracy closer to the human performance. Overall, MMM has achieved a new SOTA, i.e., test accuracy of 88.9\%, which exceeds the previous best by 16.9\%.

We also test our method on three other MCQA datasets: MCTest including MC160 and MC500, TOEFL, and SemEval-2018 Task 11. The results are summarized in Table~\ref{table:results-other-datasets}. Similarly, we list the previous SOTA models with their scores for comparison. We compared our method with the baselines that use the same sentence encoder. Except for the SemEval dataset, our method can improve the BERT-Large model by at least 10\%. For both MCTest and SemEval datasets, our best scores are very close to the reported human performance.
The MC160 and MC500 datasets were curated in almost the same way \cite{richardson2013mctest} with only one difference that MC160 is around three times smaller than MC500. We can see from Table~\ref{table:results-other-datasets} that both the BERT and RoBERTa baselines perform much worse on MC160 than MC500. We think the reason is that the data size of MC160 is not enough to well fine-tune the large models with a huge amount of trainable parameters. However, by leveraging the transfer learning techniques we proposed, we can significantly improve the generalization capability of BERT and RoBERTa models on the small datasets so that the best performance of MC160 can even surpass that of MC500. This demonstrates the effectiveness of our method.

\begin{table}
\centering
\resizebox{1.\columnwidth}{!}{\begin{tabular}{lrrr}
\Xhline{2\arrayrulewidth}
\textbf{Settings}                     & \textbf{DREAM} & \textbf{MC160} & \textbf{MC500} \\ \hline
Full Model                   &  \textbf{72.6}     &     \textbf{86.7}    & \textbf{83.5}     \\ \hline\hline
\ \ \ \ -- Second-Stage Multi-task Learning       &   68.5    &   72.5  &  78.0        \\ 
\ \ \ \ -- First-Stage Coarse-tuning on NLI &  69.5     &    80.8     & 81.8     \\ 
\ \ \ \ -- MAN                       &  71.2  &   85.4   & 81.5        \\ 
\ \ \ \ -- Speaker Normalization     &  71.4     &  ---   & ---       \\ 
\Xhline{2\arrayrulewidth}
\end{tabular}}
\caption{\small Ablation study on the DREAM and MCTest-MC160 (MC160) datasets. Accuracy (\%) is on the development set.}
\label{table:ablation}
\end{table}

To better understand why MMM can be successful, we conducted an ablation study be removing one feature at a time on the BERT-Base model. The results are shown in Table~\ref{table:ablation}. We see that the removal of the second stage multi-task learning part hurts our method most significantly, indicating that the majority of improvement is coming from the knowledge transferred from the in-domain dataset. The first stage of coarse-tuning using NLI datasets is also very important, which provides the model with enhanced language inference ability. As for the top-level classifier, i.e., the MAN module, if we replace it with a typical two-layer FCNN as in \cite{devlin-etal-2019-bert}, we have 1--2\% performance drop. Lastly, for the DREAM dataset, the speaker normalization strategy gives us another $\sim$1\% improvement.

\section{Discussion}
\label{sec:discussion}
\subsection{Why does natural language inference help?}

As shown in Table~\ref{table:ablation}, coarse-tuning on NLI tasks can help improve the performance of MCQA. We conjecture one of the reasons is that, in order to pick the correct answer, we need to rely on the language inference capability in many cases. As an example in Table~\ref{table: example}, the utterance highlighted in the bold and italic font in the dialogue is the evidence sentence from which we can obtain the correct answer to Question 2\@. There is no token overlap between the evidence sentence and the correct answer, indicating that the model cannot solve this question by surface matching. Nevertheless, the correct answer is an \textit{entailment} to the evidence sentence while the wrong answers are not. Therefore, the capability of language inference enables the model to correctly predict the answer. On the other hand, we can deem the passage and the pair of (question, answer) as a pair of \textit{premise} and \textit{hypothesis}. Then the process of choosing the right answer to a certain question is similar to the process of choosing the \textit{hypothesis} that can best entail the \textit{premise}. In this sense, the part of MCQA task can be deemed as a NLI task. This also agrees with the argument that NLI is a fundamental ability of a natural language processing model and it can help support other tasks that require higher level of language processing abilities~\cite{DBLP:journals/corr/abs-1811-00671}. We provided several more examples that require language inference reading skills in the Appendix \ref{appendix:section-2}; they are wrongly predicted by the BERT-Base baseline model but can be correctly solved by exposing the model to NLI data with the coarse-tuning stage.

\subsection{Can other tasks help with MCQA?}



By analyzing the MCQA datasets, we found that some questions ask about the attitude of one person towards something and in some cases, the correct answer is simply a paraphrase of the evidence sentence in the passage. This finding naturally leads to the question: could other kinds of tasks such as sentiment classification, paraphrasing also help with MCQA problems?

To answer this question, we select several representative datasets for five categories as the up-stream tasks: sentiment analysis, paraphrase, span-based QA, NLI, and MCQA. We conduct experiments where we first train the BERT-Base models on each of the five categories and then further fine-tune our models on the target dataset: DREAM and MC500 (MCTest-MC500). For the sentiment analysis category, we used the Stanford Sentiment Treebank (SST-2) dataset from the GLUE benchmark \cite{wang2018glue} (around 60k train examples) and the Yelp dataset\footnote{https://www.yelp.com/dataset/challenge} (around 430k train examples). For the paraphrase category, three paraphrasing datasets are used from the GLUE benchmark: Microsoft Research Paraphrase Corpus (MRPC), Semantic Textual Similarity Benchmark (STS-B), and Quora Question Pairs (QQP), which are denoted as ``GLUE-Para.''. For the span-based QA, we use the SQuAD 1.1, SQuAD 2.0
, and MRQA\footnote{https://mrqa.github.io/}
which is a joint dataset including six popular span-based QA datasets. 

Table~\ref{table:transfer-single} summarizes the results. We see that sentiment analysis datasets do not help much with our target MCQA datasets. But the paraphrase datasets do bring some improvements for MCQA.
For span-based QA, only SQuAD 2.0 helps to improve the performance of the target dataset. Interestingly, although MRQA is much larger than other QA datasets (at least six times larger), it makes the performance worst. This suggests that span-based QA might not the appropriate source tasks for transfer learning for MCQA. We hypothesis this could due to the fact that most of the questions are non-extractive (e.g., 84\% of questions in DREAM are non-extractive) while all answers are extractive in the span-based QA datasets.

For the completeness of our experiments, we also used various NLI datasets: MultiNLI, SNLI, Question NLI (QLI), Recognizing Textual Entailment (RTE), and Winograd NLI (WNLI) from the GLUE benchmark. We used them in three kinds of combinations: MultiNLI alone, MultiNLI plus SNLI denoted as ``NLI'', and combining all five datasets together, denoted as ``GLUE-NLI''. As the results shown in Table~\ref{table:transfer-single}, NLI and GLUE-NLI are comparable and both can improve the target dataset by a large margin.

Lastly, among all these tasks, using the MCQA task itself, i.e., pretraining on RACE dataset, can help boost the performance, most. This result agrees with the intuition that the in-domain dataset can be the most ideal data for transfer learning.

In conclusion, we find that for out-of-domain datasets, the NLI datasets can be most helpful to the MCQA task, indicating that the natural language inference capability should be an important foundation of the MCQA systems. Besides, a larger in-domain dataset, i.e. another MCQA dataset, can also be very useful.

\begin{table}
\centering
\small
\begin{tabular}{llrr}
\Xhline{2\arrayrulewidth}
\textbf{Task Type}                   & \textbf{Dataset Name} & \textbf{DREAM} & \textbf{MC500} \\ \hline
-                               & Baseline     & 63.2  & 69.5  \\ \hline
\multirow{2}{*}{Sentiment Analy.} & SST-2 & 62.7 & 69.5 \\
                                    & Yelp  &   62.5   & 71.0 \\ \hline
Paraphrase & GLUE-Para. & 64.2 & 72.5 \\\hline
\multirow{3}{*}{Span-based QA} & SQuAD 1.1    & 62.1  & 69.5  \\  
                               & SQuAD 2.0    & 64.0  & 74.0  \\ 
                               & MRQA         & 61.2  & 68.3  \\ \hline
\multirow{3}{*}{NLI}           & MultiNLI         & 67.0  &  79.5 \\ 
                               & NLI      & 68.4   & \underline{80.0} \\ 
                               & GLUE-NLI     & \underline{68.6} & 79.0  \\ \hline
Combination                 & GLUE-Para.+NLI & 68.0 & 79.5 \\ \hline
Multi-choice QA                & RACE         & \textbf{70.2}   & \textbf{81.2} \\ \Xhline{2\arrayrulewidth}
\end{tabular}
\caption{\small Transfer learning results for DREAM and MC500. The BERT-Base model is first fine-tuned on each source dataset and then further fine-tuned on the target dataset. Accuracy is on the the development set. A two-layer FCNN is used as the classifier.}
\label{table:transfer-single}
\end{table}




\subsection{NLI dataset helps with convergence}

\begin{figure}
    \centering
    \includegraphics[width=0.44\textwidth]{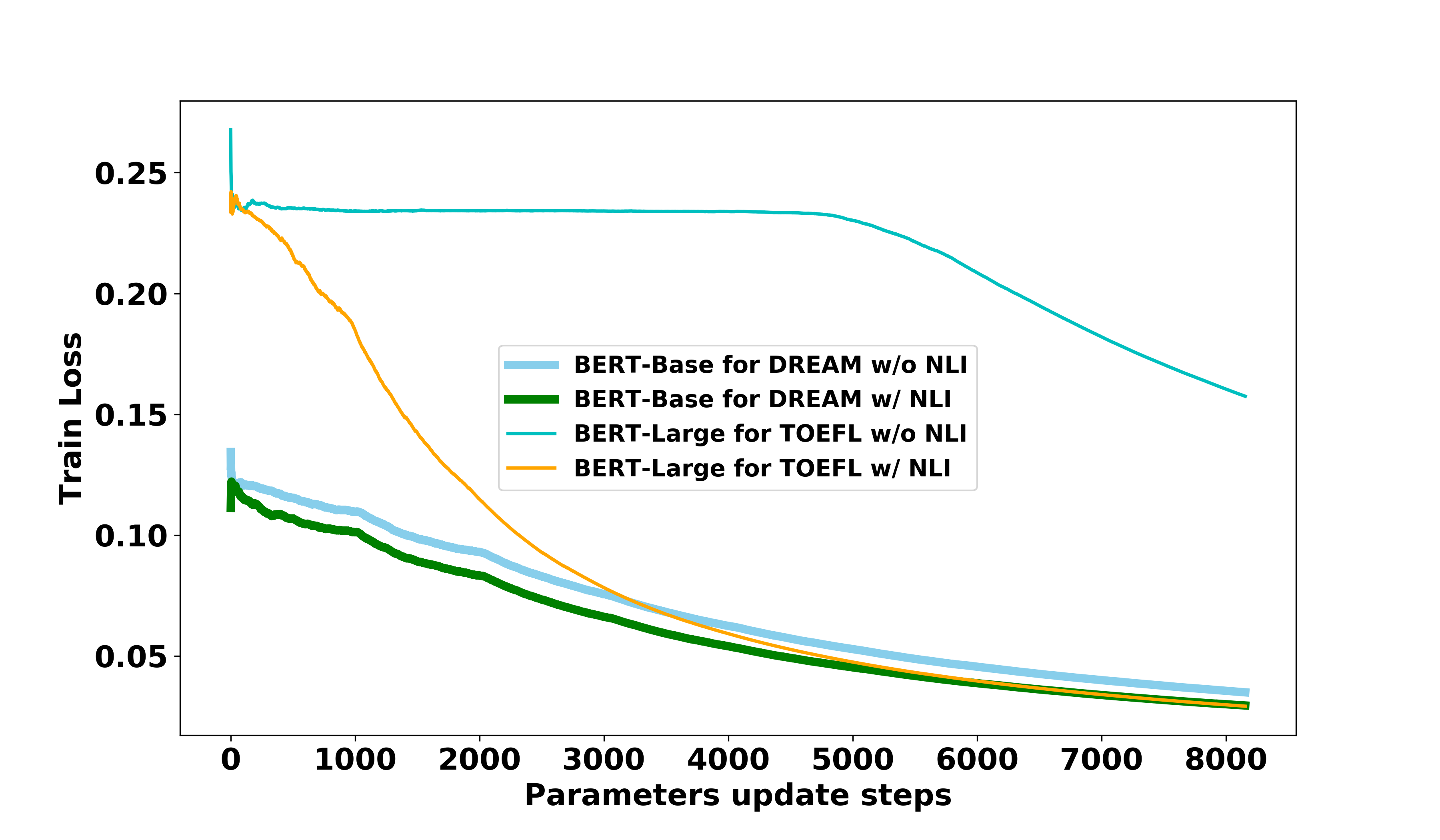}
    \caption{\small Train loss curve with respect to optimization steps. With prior coarse-tuning on NLI data, convergence becomes much faster and easier.}
    \label{fig:convergence}
\end{figure}

The first stage of coarse-tuning with NLI data can not only improve the accuracy but also help the model converge faster and better. Especially for the BERT-Large and RoBERTa-Large models that have much larger amount of trainable parameters, convergence is very sensitive to the optimization settings.
However, with the help of NLI datasets
, convergence for large models is no longer an issue, as shown in Figure~\ref{fig:convergence}. Under the same optimization hyper-parameters, compared with the baseline, coarse-tuning can make the training loss of the BERT-Base model decrease much faster. More importantly, for the BERT-Large model, without coarse-tuning, the model does not converge at all at the first several epochs, which can be completely resolved by the help of NLI data. 

\subsection{Multi-stage or Multi-task}

In a typical scenario where we have one source and one target dataset, we naturally have a question about whether we should simultaneously train a model on them via multi-task learning or first train on the source dataset then on the target sequentially. Many previous works adopted the latter way \cite{sun2018improving,chung2017supervised,phang2018sentence} and \cite{chung2017supervised} demonstrated that the sequential fine-tuning approach outperforms the multi-task learning setting in their experiments. However, we had contradictory observations in our experiments. Specifically, we conducted a pair of control experiments: one is that we first fine-tune the BERT-Base model on the source dataset RACE and then further fine-tune on the target dataset, and the other is that we simultaneously train the model on RACE and the target dataset via multi-task learning. The comparison results are shown in Table~\ref{table: multi-task}. We see that compared with sequential fine-tuning, the multi-task learning achieved better performance. We conjecture that in the sequential fine-tuning setting, while the model is being fine-tuned on the target dataset, some information or knowledge learned from the source dataset may be lost since the model is no longer exposed to the source dataset in this stage. In comparison, this information can be kept in the multi-task learning setting and thus can better help improve the target dataset.

Now that the multi-task learning approach outperforms the sequential fine-tuning setting, we naturally arrive at another question: what if we merged the coarse-tuning and multi-task learning stages together? That is, what if we simultaneously trained the NLI, source, and target datasets altogether under the multi-task learning framework? We also conducted a pair of control experiments for investigation. The results in Table~\ref{table: multi-task},
show that casting the fine-tuning process on these three datasets into separate stages performs better, indicating that multi-stage training is also necessary. 
Considering that the NLI dataset is an out-of-domain dataset while RACE is in-domain with respect to the target datasets, we can obtain a good practice: we first separate the source datasets into two categories: out-of-domain and in-domain, based on the type of the target dataset; then we can adopt a multi-stage training strategy, that is, first fine-tune the model on the out-of-domain source datasets, then fine-tune on the in-domain source datasets and the target dataset together via multi-task learning.

\begin{table}
\centering
\small
\resizebox{1.\columnwidth}{!}{\begin{tabular}{lrrr}
\Xhline{2\arrayrulewidth}
 \textbf{Setting Configuration}& \textbf{DREAM} & \textbf{MC160} & \textbf{MC500} \\ \hline
 BERT-Base -\textgreater RACE -\textgreater Target                    & 70.2    & 80.0
       &  81.2   \\ 
BERT-Base -\textgreater \{RACE, Target\}                             & \textbf{70.7} &   \textbf{80.8} &  \textbf{81.8}    \\ \hline
 BERT-Base -\textgreater \{RACE, Target, NLI\} & 70.5   &   87.0      &  82.5   \\ 
BERT-Base -\textgreater NLI -\textgreater \{RACE, Target\}          & \textbf{71.2}  & \textbf{88.3} &  \textbf{83.5}   \\ \Xhline{2\arrayrulewidth}
\end{tabular}
}
\caption{\small Comparison between multi-task learning and sequential fine-tuning. BERT-Base model is used and the accuracy is on the development set. Target refers to the target dataset in transfer learning. A two-layer FCNN instead of MAN is used as the classifier.}
\label{table: multi-task}
\end{table}




\subsection{Multi-steps reasoning is important}

\begin{figure}
    \small
    \centering
    \includegraphics[width=0.34\textwidth]{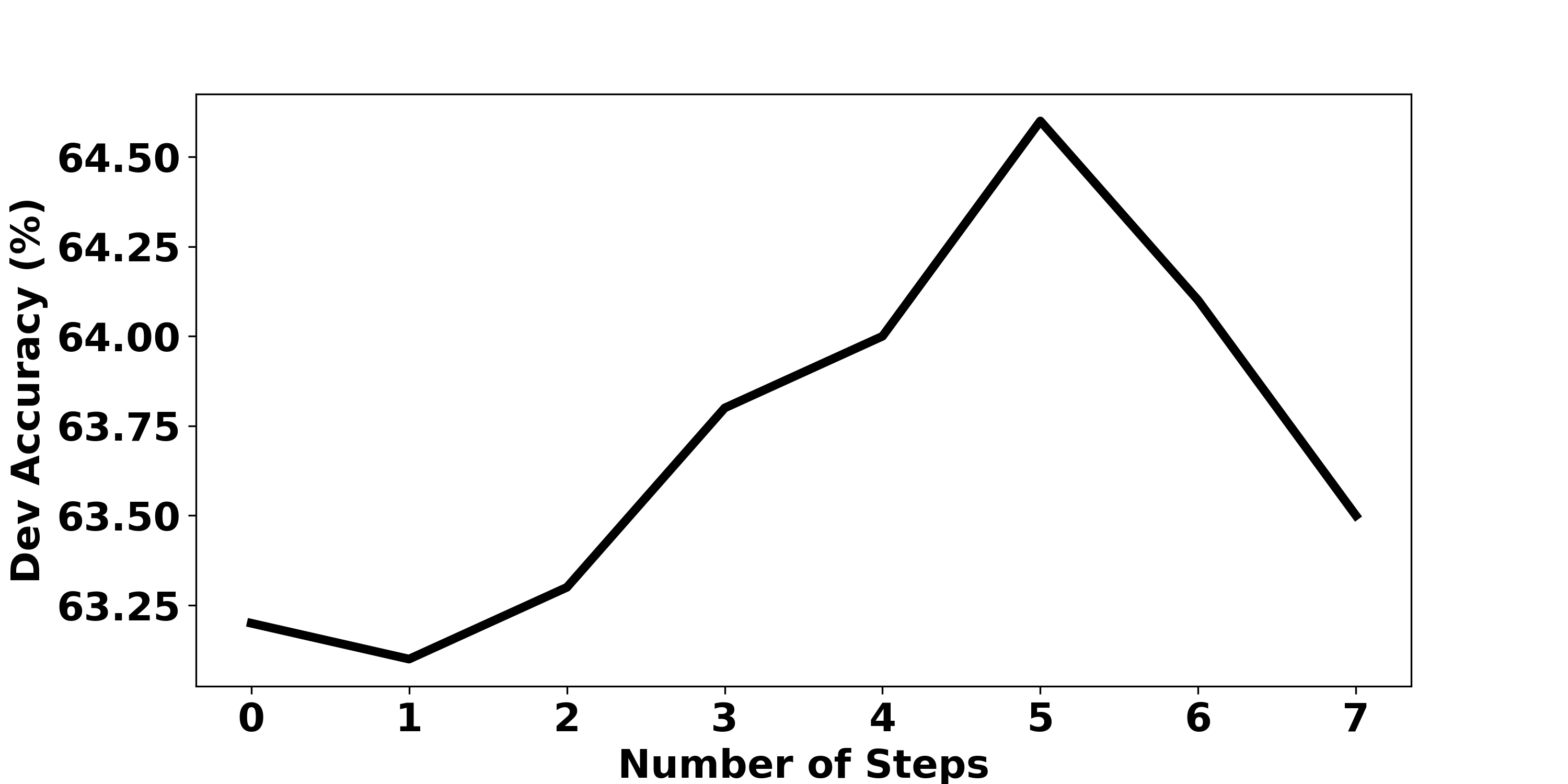}
    \caption{\small Effects of the number of reasoning steps for the MAN classifier. 0 steps means using FCNN instead of MAN. The BERT-Base model and DREAM dataset are used.}
    \label{fig:MAN}
\end{figure}

Previous results show that the MAN classifier shows improvement compared with the FCNN classifier, but we are also interested in how the performance change while varying the number of reasoning steps $K$ as shown in Figure~\ref{fig:MAN}. $K=0$ means that we do not use MAN but FCNN as the classifier. We observe that there is a gradual improvement as we increase $K=1$ to $K=5$, but after 5 steps the improvements have saturated. This verifies that an appropriate number of steps of reasoning is important for the memory network to reflect its benefits. 

\subsection{Could the source dataset be benefited?}

So far we have been discussing the case where we do multi-task learning with the source dataset RACE and various much smaller target datasets to help improve the targets. We also want to see whether our proposed techniques can also benefit the source dataset itself. Table~\ref{table:race-ablation} summarizes the results of BERT-Base model on the RACE dataset obtained by adding the coarse-tuning stage, adding the multi-task training together with DREAM, and adding the MAN module. From this table, we see that all three techniques can bring in improvements over the baseline model for the source dataset RACE, among which NLI coarse-tuning stage can help elevate the scores most. 

Since we found all parts of MMM can work well for the source dataset, we tried to use them to improve the accuracy on RACE. The results are shown in Table~\ref{table:race-best}. We used four kinds of pre-trained sentence encoders: BERT-Base, BERT-Large, XLNet-Large, and RoBERTa-Large. For each encoder, we listed the official report of scores from the leaderboard.
Compared with the baselines, MMM leads to improvements ranging from 0.5\% to 3.0\% in accuracy. Our best result is obtained by the RoBERTa-Large encoder.

\begin{table}
\centering
\small
\begin{tabular}{lrrr}
\Xhline{2\arrayrulewidth}
\textbf{Settings}    & \textbf{RACE-M} & \textbf{RACE-H} & \textbf{RACE} \\ \hline
BERT-Base    &   73.3     &  64.3      &   66.9   \\ 
\ \ \ \ +NLI     &  \textbf{74.2} &   \textbf{66.6}     &   \textbf{68.9}   \\ 
\ \ \ \ +DREAM &  72.4      &   66.1     &   67.9   \\ 
\ \ \ \ +MAN        &  71.2      &   \textbf{66.6}     &   67.9   \\ \Xhline{2\arrayrulewidth}
\end{tabular}
\caption{\small Ablation study for the RACE dataset. The accuracy is on the development set. All parts of MMM improve this source dataset.}
\label{table:race-ablation}
\end{table}

\begin{table}
\small
\centering
\resizebox{.95\columnwidth}{!}{\begin{tabular}{lrrr}
\Xhline{2\arrayrulewidth}
\textbf{Model}         & \textbf{RACE-M} & \textbf{RACE-H} & \textbf{RACE} \\ \hline
\textit{Official Reports:} & & & \\
BERT-Base     &   71.7	  &  62.3      &   65.0   \\ 
BERT-Large    &   76.6     &  70.1      &   72.0   \\ 
XLNet-Large   &   85.5     &  80.2      &  81.8    \\ 
RoBERTa-Large &   86.5     &    81.3    &  83.2    \\ \hline\hline
BERT-Base+MMM     &   74.8     &  65.2      &   68.0   \\ 
BERT-Large+MMM    &   78.1     &  70.2      &   72.5   \\ 
XLNet-Large+MMM   &   86.8     &  81.0      &  82.7    \\ 
RoBERTa-Large+MMM &   \textbf{89.1}     &  \textbf{83.3}      &   \textbf{85.0}   \\ \Xhline{2\arrayrulewidth}
\end{tabular}}
\caption{\small Comparison of the test accuracy of the RACE dataset between our approach MMM and the official reports that are from the dataset leaderboard.}
\label{table:race-best}
\end{table}

\subsection{Error Analysis}

\begin{table}
\centering
\resizebox{.9\columnwidth}{!}{\begin{tabular}{llrr}
\Xhline{2\arrayrulewidth}
\textbf{Major Types}                & \textbf{Sub-types}    & \textbf{Percent} & \textbf{Accuracy} \\ \hline
\multirow{2}{*}{Matching}  & Keywords     & 23.3    & 94.3     \\  
                           & Paraphrase   & 30.7    & 84.8     \\ \hline
\multirow{3}{*}{Reasoning} & Arithmetic   & 12.7    & 73.7     \\  
                           & Common Sense & 10.0    & 60.0     \\  
                           & Others       & 23.3    & 77.8     \\ \Xhline{2\arrayrulewidth}
\end{tabular}}
\caption{\small Error analysis on DREAM. The column of ``Percent'' reports the percentage of question types among 150 samples that are from the development set of DREAM dataset that are wrongly predicted by the BERT-Base baseline model. The column of ``Accuracy'' reports the accuracy of our best model (RoBERTa-Large+MMM) on these samples.}
\label{table:error}
\end{table}

In order to investigate how well our model performs for different types of questions, we did an error analysis by first randomly selecting 150 samples from the common wrong predictions on the development set of DREAM dataset, obtained by three BERT-Base baseline models, each of which was individually trained with different random seeds. We then manually classified them into several question types, as shown in Table~\ref{table:error}. The annotation criterion is described in the Appendix \ref{appendix:section-3}. We see that the BERT-Base baseline model still does not do well on matching problems.
 We then evaluate our best model on these samples and report the accuracy of each question type in the last column of Table~\ref{table:error}. We find that our best model can improve upon every question type significantly especially for the matching problems, and most surprisingly, our best model can even greatly improve its ability on solving the arithmetic problems, achieving the accuracy of 73.7\%.
 
 However, could our model really do math? To investigate this question, we sampled some arithmetic questions that are correctly predicted by our model, made small alterations to the passage or question, and then checked whether our model can still make correct choices. We found our model is very fragile to these minor alterations, implicating that the model is actually not that good at arithmetic problems. We provided one interesting example in the Appendix \ref{appendix:section-3}.

\section{Related Work}


There are increasing interests in  machine reading comprehension (MRC) for question answering (QA). The extractive QA tasks primarily focus on locating text spans from the given document/corpus to answer questions \cite{rajpurkar2018know}. Answers in abstractive datasets such as MS MARCO \cite{nguyen2016ms}, SearchQA \cite{dunn2017searchqa}, and NarrativeQA \cite{kovcisky2018narrativeqa} are human-generated and based on source documents or summaries in free text format. However, since annotators tend to copy spans as answers \cite{reddy2019coqa}, the majority of answers are still extractive in these datasets. The multi-choice QA datasets are collected either via crowd sourcing, or collected from examinations designed by educational experts \cite{lai2017race}. In this type of QA datasets, besides token matching, a significant portion of questions require multi-sentence reasoning and external knowledge \cite{ostermann2018semeval}. 

Progress of research for MRC first relies on the breakthrough of the sentence encoder, from the basic LSTM to the pre-trained transformer based model \cite{devlin-etal-2019-bert}, which has elevated the performance of all MRC models by a large margin. Besides, the attention mechanisms between the context and the query can empower the neural models with higher performance \cite{seo2016bidirectional}. 
In addition, some techniques such as answer verification \cite{hu2019read+}, multi-hop reasoning \cite{xiao2019dynamically}, and synthetic data augmentation can be also helpful.


Transfer learning has been widely proved to be effective across many domain in NLP. In the QA domain, the most well-known example of transfer learning would be fine-tuning the pre-trained language model such as BERT to the down-streaming QA datasets such as SQuAD \cite{devlin-etal-2019-bert}. Besides, multi-task learning can also be deemed as a type of transfer learning, since during the training of multiple datasets from different domains for different tasks, knowledge will be shared and transferred from each task to others, which has been used to build a generalized QA model \cite{talmor2019multiqa}. However, no previous works have investigated that the knowledge from the NLI datasets can also be transferred to improve the MCQA task.



\section{Conclusions}

We propose MMM, a multi-stage multi-task transfer learning method on the multiple-choice question answering tasks. Our two-stage training strategy and the multi-step attention network achieved significant improvements for MCQA. We also did detailed analysis to explore the importance of both our training strategies as well as different kinds of in-domain and out-of-domain datasets. 
It is noteworthy that our proposal transfer learning strategy can actually be generalized to various NLP tasks, where for any given target dataset, we can find its corresponding out-of-domain and in-domain source datasets, and then we train the model on the out-of-domain source datasets first, and subsequently fine-tune the model on the combination of the in-domain datasets and the target datasets via multi-task training. This strategy should always be effective at improving the target dataset.


\bibliographystyle{aaai}
\bibliography{references}

\newpage
\appendix
\section{Appendix}
\setcounter{table}{0}
\renewcommand{\thetable}{A\arabic{table}}

\subsection{Optimization Hyper-parameters}\label{appendix:section-1}

The learning rate and number of training epochs vary for different datasets and encoder types, which are summarized in Table \ref{table:lr} and \ref{table:epochs}, respectively, for references.

\begin{table}[!ht]
\centering
\small
\begin{tabular}{lccc}
\Xhline{2\arrayrulewidth}
 \textbf{Datasets}            & \textbf{BERT-Base} & \textbf{BERT-Large} & \textbf{RoBERTa-Large} \\ \hline
\textbf{DREAM}        & 2e-5      & 2e-5       & 1e-5          \\
\textbf{MCTest}       & 1e-5      & 5e-6       & 5e-6          \\
\textbf{TOEFL}        & 5e-6      & 1e-5       & 5e-6          \\
\textbf{SemEval} & 2e-5      & 2e-5       & 1e-5          \\
\textbf{RACE}         & 5e-5      & 2e-5       & 1e-5         \\
\Xhline{2\arrayrulewidth}
\end{tabular}
\caption{Optimal learning rate for different datasets and encoder types.}
\label{table:lr}
\end{table}

\begin{table}[!ht]
\centering
\small
\begin{tabular}{lccc}
\Xhline{2\arrayrulewidth}
  \textbf{Datasets}           & \textbf{BERT-Base} & \textbf{BERT-Large} & \textbf{RoBERTa-Large} \\ \hline
\textbf{DREAM}        & 8         & 8          & 10            \\
\textbf{MCTest}       & 8         & 8          & 10            \\
\textbf{TOEFL}        & 8         & 8          & 10            \\
\textbf{SemEval} & 8         & 8          & 10            \\
\textbf{RACE}         & 5         & 5          & 10           \\
\Xhline{2\arrayrulewidth}
\end{tabular}
\caption{Optimal number of training epochs for different datasets and encoder types.}
\label{table:epochs}
\end{table}

\subsection{Natural Inference Helps Making Choices}\label{appendix:section-2}

Now that exposing the model to Natural Language Inference (NLI) data can help improve its performance in the multi-choice question answering (MCQA) task, we showcase several examples that are wrongly predicted by the BERT-Base baseline model but are correctly solved by incorporating the first stage of coarse-tuning with NLI data in the Table \ref{table:examples}. Exposing the model to NLI data can help enhance its language inference ability, which is required by all these examples to get correct answers.

\begin{table}[!ht]
\centering
\small
\resizebox{\columnwidth}{!}{\begin{tabular}{l}
\Xhline{2\arrayrulewidth}
\textbf{Dialogue 1:}                                                               \\
man:  Wonderful day, isn't it? Want to join me for a swim? \\
woman:  \textbf{\textit{If you don't mind waiting while I get prepared.}} \\ \hline
\textbf{Question}: What does the woman mean?                                       \\ 
\textbf{A.} She is too busy to go.                                               \\ 
\textbf{B.} She doesn't want to wait long. $\times$                                                 \\ 
\textbf{C.} She's willing to go swimming. $\surd$                                            \\ 
\hline \hline
\textbf{Dialogue 2:}                                                               \\ 
woman:  Shall we go to a play or to a movie? \\
man:  \textbf{\textit{It's all the same to me.}} \\ \hline
\textbf{Question}: What does this man mean?                                       \\ 
\textbf{A.} It makes no difference to him which they go to. $\surd$                                               \\ 
\textbf{B.} He does not want to go to either one. $\times$                                                 \\ 
\textbf{C.} The play and the movie are about the same subject.                                             \\
\hline \hline
\textbf{Dialogue 3:}                                                               \\ 
woman:  I'm sorry, Mr Wilson. I got up early but the bus was late. \\
man:  \textbf{\textit{Your bus is always late, Jane.}} \\ \hline
\textbf{Question}: What does the man mean?                                       \\ 
\textbf{A.} Jane used the same excuse again. $\surd$                                               \\ 
\textbf{B.} Jane stayed up too late last night.                                                  \\ 
\textbf{C.} Jane always gets up early. $\times$                                             \\
\Xhline{2\arrayrulewidth}
\end{tabular}}
\caption{Examples from the DREAM dataset. $\surd$ marks the correct answer and the answer chosen by the NLI data enhanced BERT-Base model while $\times$ marks the answer predicted by the BERT-Base baseline model. These examples are wrongly solved by the BERT-Base baseline model but get correct predictions by inserting the first stage of coarse-tuning using NLI data.}
\label{table:examples}
\end{table}

\subsection{Error Analysis}\label{appendix:section-3}

To conduct the error analysis, we randomly selected 150 samples that are from the development set of DREAM dataset and wrongly predicted by the BERT-Base baseline model. We then manually classified them into several question types based on the following criterion:

\begin{itemize}
    \item \textbf{Matching:}
    \begin{itemize}
        \item \textbf{Keywords:} The correct answer is a phrase and can match a span of text in the passage.
        \item \textbf{Paraphrase:} The correct answer is a sentence and is a paraphrase to the evidence sentence in the passage.
    \end{itemize}
    \item \textbf{Reasoning:}
    \begin{itemize}
        \item \textbf{Arithmetic:} The correct answer is a number and some calculations must be conducted to get it.
        \item \textbf{Common Sense:} Some common sense is needed to answer the question.
        \item \textbf{Others:} Other kinds of questions that need some reasoning to obtain the answer.
    \end{itemize}
\end{itemize}

By evaluating the accuracy of our best model on each of these question types, we found our model can even do very well on the arithmetic problems. In order to verify whether our model really has the ability of doing math, we sampled some arithmetic questions that are correctly predicted by our model, made small alterations to the passage or (question, answer) pair, and then checked whether our model can still make correct choices. Table \ref{table:error-example-1} shows one arithmetic problem and our model can get it right. The correct answer ``86 dollars'' should be the addition of the car rent ``78 dollars'' and the car insurance ``8 dollars'', and it seems that our model can perform this simple calculation. However, if we simply changed the car insurance price from 8 dollars to 7 dollars in the passage, the model would obtain the wrong prediction ``71 dollars'' as shown in Table \ref{table:error-example-2}. We also curated another type of adversarial example by revising the passage so that the woman in the dialogue does not want the car insurance, in which the correct answer should be only the car rent price ``78 dolllars''. As shown in Table \ref{table:error-example-3}, our model again makes the wrong choice ``70 dollars''. These two adversarial examples strongly disprove that our model really has the ability of solving mathematical questions.

\begin{table}[!ht]

\begin{subtable}{1\columnwidth}
\centering
\small
\resizebox{\columnwidth}{!}{\begin{tabular}{l}
\Xhline{2\arrayrulewidth}
\textbf{Dialogue:}                                                               \\
man:  Good morning. May I help yon? \\
woman:  I'd like to rent a car, please. \\
man:  Okay. Full-size, mid-size, or compact, madam? \\
woman:  Compact is OK. What's the rate? \\
man:  78 dollars a day. \\
woman:  And I'd like to have insurance just in case. \\
man:  If you want full coverage insurance, it will be 8 dollars per day. \\
woman:  All right, I'll take that, too. \\
man: OK. Please fill in this form. \\
\hline
\textbf{Question:} How much will the woman pay in total?                                       \\ 
\textbf{A.} 70 dollars.                                               \\ 
\textbf{B.} 78 dollars.                                                  \\ 
\textbf{C.} 86 dollars. $\surd$                                            \\ \hline
\textbf{Model Prediction:} C. 86 dollars. \\
\Xhline{2\arrayrulewidth}
\end{tabular}}
\caption{Example of arithmetic question correctly solved by our best model. This example is from the development set of the DREAM dataset. $\surd$ marks the correct answer.}
\label{table:error-example-1}
\end{subtable}

\bigskip
\begin{subtable}{1\columnwidth}
\centering
\small
\resizebox{\columnwidth}{!}{\begin{tabular}{l}
\Xhline{2\arrayrulewidth}
\textbf{Dialogue:}                                                               \\
man:  Good morning. May I help yon? \\
woman:  I'd like to rent a car, please. \\
man:  Okay. Full-size, mid-size, or compact, madam? \\
woman:  Compact is OK. What's the rate? \\
man:  78 dollars a day. \\
woman:  And I'd like to have insurance just in case. \\
man:  If you want full coverage insurance, it will be \textbf{\textit{7}} dollars per day. \\
woman:  All right, I'll take that, too. \\
man: OK. Please fill in this form. \\
\hline
\textbf{Question:} How much will the woman pay in total?                                       \\ 
\textbf{A.} \textbf{\textit{71}} dollars.                                               \\ 
\textbf{B.} 78 dollars.                                                  \\ 
\textbf{C.} \textbf{\textit{85}} dollars. $\surd$                                            \\ \hline
\textbf{Model Prediction:} A. 71 dollars. \\
\Xhline{2\arrayrulewidth}
\end{tabular}}
\caption{Adversarial example that forces our best model to make wrong predictions and is crafted by slightly revising the example in Table \ref{table:error-example-1}. The revisions are highlighted in bold and italic font. $\surd$ marks the correct answer.}
\label{table:error-example-2}
\end{subtable}

\bigskip
\begin{subtable}{1\columnwidth}
\centering
\small
\resizebox{\columnwidth}{!}{\begin{tabular}{l}
\Xhline{2\arrayrulewidth}
\textbf{Dialogue:}                                                               \\
man:  Good morning. May I help yon? \\
woman:  I'd like to rent a car, please. \\
man:  Okay. Full-size, mid-size, or compact, madam? \\
woman:  Compact is OK. What's the rate? \\
man:  78 dollars a day. \\
woman:  And I'd like to have insurance just in case. \\
man:  If you want full coverage insurance, it will be 7 dollars per day. \\
woman:  \textbf{\textit{Oh, that's too expansive for me. Then I would rather not}} \\ \textbf{\textit{ to have the insurance.}} \\
man: \textbf{\textit{All right, I will cancel that for you.}} Please fill in this form. \\
\hline
\textbf{Question:} How much will the woman pay in total?                                       \\ 
\textbf{A.} 70 dollars.                                               \\ 
\textbf{B.} 78 dollars. $\surd$                                                  \\ 
\textbf{C.} 86 dollars.                                             \\ \hline
\textbf{Model Prediction:} A. 70 dollars. \\
\Xhline{2\arrayrulewidth}
\end{tabular}}
\caption{Adversarial example that forces our best model to make wrong predictions and is crafted by slightly revising the example in Table \ref{table:error-example-1}. The revisions are highlighted in bold and italic font. $\surd$ marks the correct answer.}
\label{table:error-example-3}
\end{subtable}
\caption{Error analysis on arithmetic problems.} 
\label{table:error-examples}
\end{table}

\end{document}